\documentclass[11pt]{article}

\usepackage{acl}
 
\usepackage{times}
\usepackage{latexsym}
\usepackage{graphicx}
\usepackage{adjustbox}
\usepackage[T1]{fontenc}

\usepackage[utf8]{inputenc}

\usepackage{microtype}

\usepackage{inconsolata}

\usepackage{booktabs}
\usepackage{colortbl}      
\usepackage{multirow}      
\usepackage{threeparttable} 
\usepackage{xcolor}
\usepackage{amsmath}
\usepackage{fancyhdr}
\usepackage{graphicx}
\fancypagestyle{firstpage}{
    \fancyhf{}                           
    \lhead{\includegraphics[width=3cm]{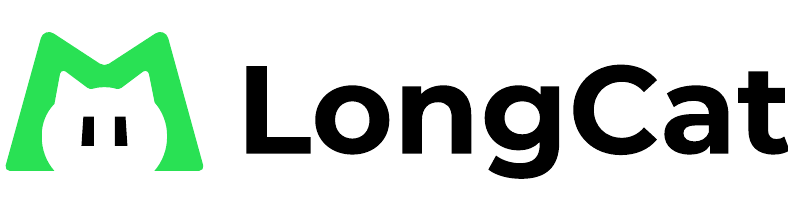}}  
    \renewcommand{\headrulewidth}{0pt}   
}

\definecolor{mygray}{rgb}{0.93, 0.93, 0.93}

\title{LoHoSearch: Benchmarking Long-Horizon Search Agents Beyond the Human Difficulty Ceiling}

\author{
 \textbf{Jiarui Zhao\textsuperscript{*$\dagger$}},
 \textbf{Rongzhi Zhang\textsuperscript{*}},
 \textbf{Lingchuan Liu\textsuperscript{$\dagger$}},
 \textbf{Hao Yang\textsuperscript{}},
\\
 \textbf{Xunliang Cai\textsuperscript{}},
 \textbf{Xi Su\textsuperscript{}}
\\
 \textsuperscript{}Meituan
\\
{\{zhaojiarui02,liulingchuan\}}@meituan.com
}
\begin{document}
\maketitle
\renewcommand{\thefootnote}{}  
\footnotetext{$^{*}$These authors contributed equally to this work.}
\footnotetext{$^\dagger$Corresponding authors.}
\renewcommand{\thefootnote}{\arabic{footnote}}  
\thispagestyle{firstpage} 
\renewcommand{\headrulewidth}{0pt}  

\begin{abstract}
Search agent benchmarks exemplified by BrowseComp have rapidly saturated over the past year, with the strongest models surpassing 90\% accuracy. Since these benchmarks are predominantly human-authored, annotators lack a global perspective on entity statistics and cannot systematically maximize search space size and structural complexity. This creates a difficulty ceiling that is hard to break. To address this, we introduce \textbf{LoHoSearch} (\textbf{Lo}ng-\textbf{Ho}rizon \textbf{Search} Agents), a challenging benchmark comprising 544 human-verified questions across 11 domains. LoHoSearch is constructed via an automated pipeline built upon a knowledge graph covering over 7  million Wikipedia entities, which selects relations with large search spaces and assembles them into structurally complex questions with KG-verified unique answers. Our evaluation demonstrates that even the strongest model achieves only 34.74\% accuracy, and existing context management strategies (best +6.8\%) yield far smaller gains than on prior benchmarks. LoHoSearch provides a more demanding standard for evaluating long-horizon reasoning and context management in search agents. The benchmark is publicly available at \url{https://huggingface.co/datasets/meituan-longcat/LoHoSearch}.
\end{abstract}

\section{Introduction}

\begin{figure}[t]
    \centering    \includegraphics[width=\linewidth]{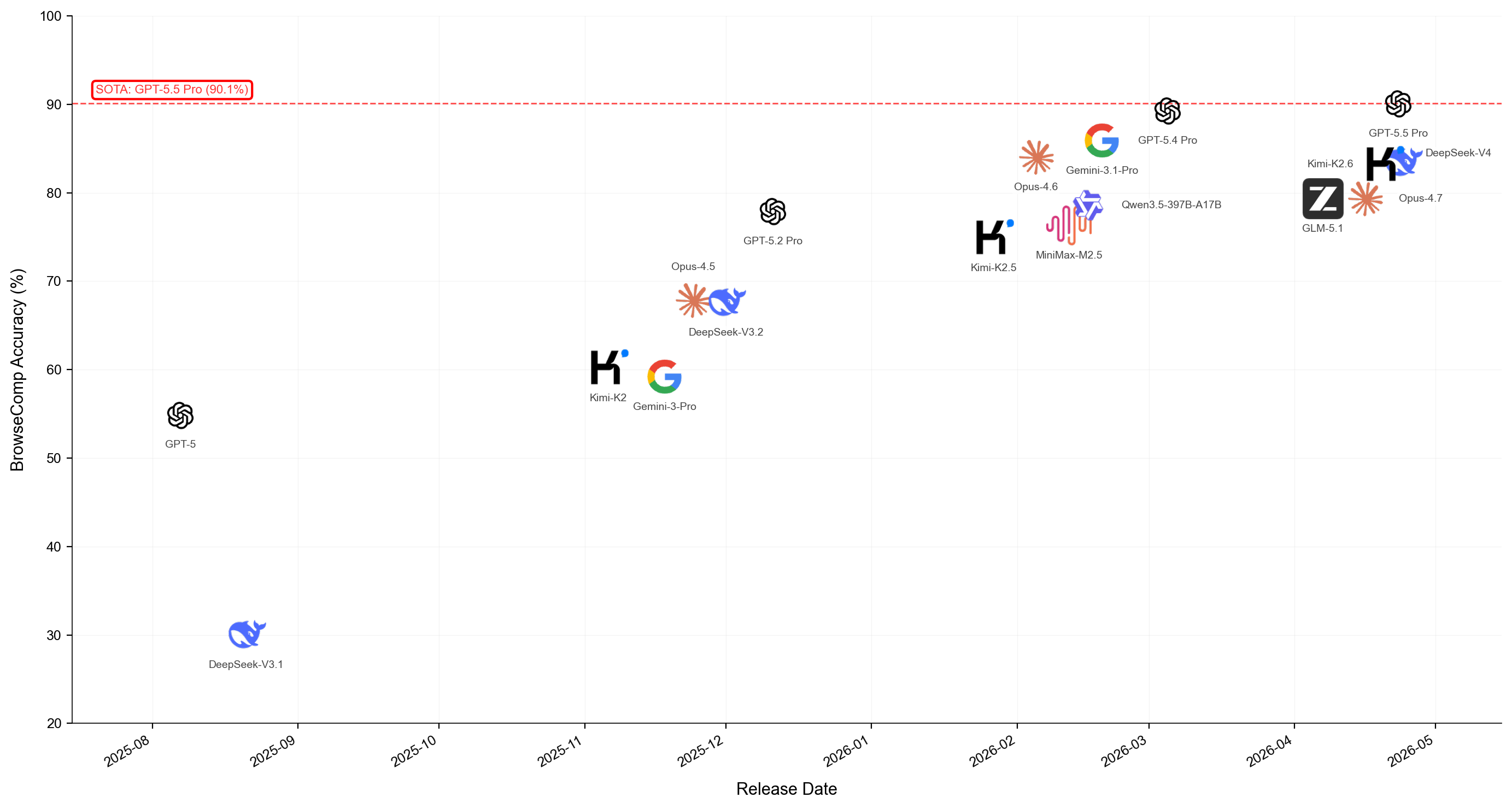}
    \caption{BrowseComp accuracy progression from August 2025 to May 2026 across major model families.}
\label{fig:browsecomp_progression}
\end{figure}

Since April 2025, challenging yet easily verifiable benchmarks, exemplified by BrowseComp~\citep{wei2025browsecomp}, have rapidly become the de facto standard for measuring search agent capabilities. Yet, as Figure~\ref{fig:browsecomp_progression} illustrates, model performance on BrowseComp has soared from 30\% to over 90\% in barely ten months, and the benchmark is quickly losing its discriminating power~\citep{anthropic2026claudemythos}. The root cause is that these benchmarks are predominantly human-authored: annotators tend to choose entities and relations they are familiar with, which typically have high popularity and direct connections, causing most questions to be answerable within only a few retrieval steps. This forms a difficulty ceiling that is hard to raise further, and as model capabilities continue to advance, this trend will only intensify.

The difficulty of search problems is determined by two core factors: (1)~the search space size per constraint, i.e., the number of candidate entities satisfying a single condition. A larger search space forces the agent to verify and eliminate more candidates. (2)~The structural complexity, i.e., the number of constraints that must be jointly satisfied to uniquely identify the answer. Higher structural complexity means more constraints must be checked to rule out each candidate, substantially raising the overall solving difficulty. While reasoning depth (the number of knowledge hops required to reach the final answer) also contributes to difficulty, it is the easiest to control and has been well-addressed by existing benchmarks~\citep{trivedi-etal-2022-musique, krishna-etal-2025-fact}. When both search space size and structural complexity are large, the agentic search process becomes substantially longer, placing higher demands on both reasoning and context management. However, human annotators lack a global perspective on entity statistics and cannot systematically maximize difficulty along both dimensions.

To address this, we introduce LoHoSearch, a benchmark constructed through an automated pipeline grounded in a knowledge graph. Starting from Wikipedia, we build a large-scale knowledge graph spanning over 7~million entities, select relations with genuinely large search spaces under a global view, and assemble them into structurally complex subgraphs whose answers are KG-verified uniqueness. Each subgraph is subsequently verbalized into a natural-language question by a language model and undergoes multiple rounds of automated verification and human review to ensure correctness and answer uniqueness. The resulting benchmark comprises 544 human-verified questions across 11 topical domains. Our main contributions are as follows:

\begin{itemize}
    \item We propose a knowledge-graph-based automated QA construction pipeline that systematically controls search space size and structural complexity, breaking through the difficulty ceiling of human authoring.
    \item We introduce the LoHoSearch benchmark. Even the strongest model achieves only 34.7\%, with correct trajectories requiring 1.7$\times$ more tool calls than BrowseComp, establishing a more discriminative evaluation standard for search agents.
    \item Our benchmark reveals the limitations of existing context management strategies in high-difficulty scenarios. The best strategy yields only a 6.8\% improvement, far below gains on existing benchmarks, demonstrating that LoHoSearch serves as a more challenging testbed for future research.
\end{itemize}

\section{Data Synthesis}

\begin{figure*}[t]
    \centering
    \includegraphics[width=\linewidth]{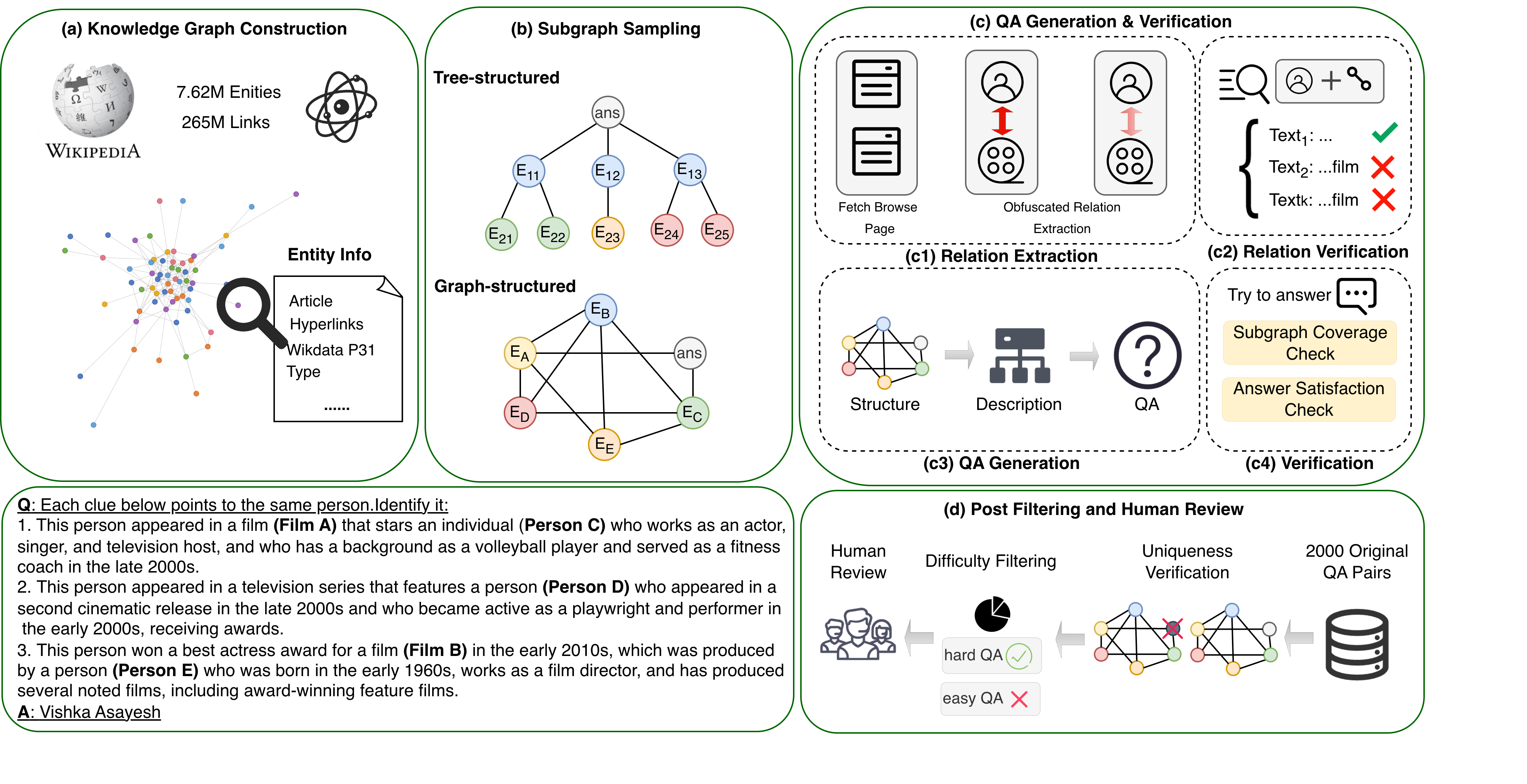}
    \caption{
        Overview of the LoHoSearch pipeline.
    }
    \label{fig:pipeline}
\end{figure*}

Our pipeline proceeds through four stages (illustrated in Figure~\ref{fig:pipeline}): knowledge graph construction, subgraph sampling, QA generation and verification, and post filtering with human review.

\subsection{Knowledge Graph Construction}
We construct a knowledge graph from the full English Wikipedia dump: each page corresponds to an entity (node), and hyperlinks within the page body pointing to other Wikipedia pages serve as directed edges. We define each entity's type as its Wikidata~\citep{wikidata2014denny} P31 (\texttt{instance\_of}) class and its popularity as in-degree, both for use in subsequent stages. The resulting knowledge graph contains approximately 7.62 million entities and 265 million directed edges.

\subsection{Subgraph Sampling}
We employ two complementary subgraph structures: tree-structured and graph-structured. The difficulty of the tree structure mainly arises from the size of the search space, while the graph structure further increases structural complexity through cyclic dependencies and cross-constraints among entities. Both structures require all constituent entities to have low popularity and moderate page lengths, ensuring that entities cannot be easily inferred. Additionally, we balance the type distribution of answer entities during sampling to ensure diversity across topical domains.

We first define the search space of a relation. Given a directed edge from entity $A$ to entity $B$, its search space is defined as:
\begin{equation}
\begin{split}
\mathcal{S}(A \!\rightarrow\! B) = \{e \mid\; & \text{type}(e) = \text{type}(A), \\
& (e \!\rightarrow\! B) \in \mathcal{G}\}
\end{split}
\end{equation}
That is, in the knowledge graph $\mathcal{G}$, the set of all entities that share the same type as $A$ and also have a directed edge to $B$. A larger search space means more candidate entities satisfy the given relational constraint, making it harder for the agent to identify $A$ via this relation.

\subsubsection{Tree-Structured Subgraph Sampling}
The tree structure uses a low-popularity entity as the root node (i.e., the answer), which connects to multiple intermediate entities, each of which further connects to several leaf nodes. Sampling proceeds layer by layer:

\paragraph{First-layer expansion.} From the root's relations, we select $N$ edges pointing to intermediate entities, subject to:
\begin{itemize}
    \item The search space size of each relation $\lvert \mathcal{S} \rvert > \tau$;
    \item The intersection of candidate sets for any $N{-}1$ relations has size $> 1$. That is, the answer is no longer uniquely determined if we remove any single relation, ensuring every relation is necessary;
    \item The intersection of candidate sets across all $N$ relations equals exactly $\{\text{root}\}$, guaranteeing KG-level uniqueness of the answer.
\end{itemize}

\paragraph{Second-layer expansion.} For each intermediate entity, we select 1 to $M$ edges pointing to leaf nodes, subject to:
\begin{itemize}
    \item The search space size of each relation $\lvert \mathcal{S} \rvert > \tau$;
    \item The intersection of candidate sets across the $M$ relations has size $> 1$, ensuring the intermediate entity itself cannot be directly inferred;
    \item We refer to candidates in this intersection (other than the current intermediate entity) as pseudo-candidates. We further require that no pseudo-candidate, when combined with the remaining intermediate entities, uniquely determines the answer—this preserves answer uniqueness.
\end{itemize}

In practice, we set $N = 3$, $M = 2$, and $\tau = 3$.

\subsubsection{Graph-Structured Subgraph Sampling}
Unlike the hierarchical expansion of tree structures, graph-structured subgraphs contain extensive cross-edges among entities and may form cycles, making the problem constraints non-decomposable into independent sub-problems. During sampling, we first select a low-popularity entity as the seed (i.e., the answer), then greedily expand until the subgraph reaches a maximum of 10 entities: at each step, we prioritize the candidate with the most edges to the current subgraph and whose corresponding edges have the largest search space. The subgraph must satisfy node type diversity, sufficient edge count, and connectivity.

After construction, uniqueness is verified via exhaustive backtracking search: we search for another set of entities in the full graph that satisfies the same entity types and directed adjacency relations; if none exists, uniqueness is confirmed. Additionally, we require the seed entity to have a sufficient number of confounding candidates of the same type that are connected to entities of all neighbor types in the subgraph, preventing brute-force solving via type enumeration.

\subsection{QA Generation and Verification}
This stage converts each sampled subgraph into a natural-language question. Specifically, for each edge in the subgraph, a language model extracts an obfuscated description of the relation from the source entity's Wikipedia page; for leaf nodes in tree structures, 1--2 additional property descriptions are extracted.

After extraction, we apply search-based verification to ensure sufficient obfuscation: each relation must be neither directly locatable via search engines nor inferable by an LLM. To prevent multiple relations from becoming easy to infer when combined, we perform joint verification on all relations of the same entity.

Once verification passes, we assemble all relations and properties into a structured description with entity names hidden, and have an LLM convert it into a natural-language question. The generated question undergoes two rounds of automated validation:
\begin{itemize}
    \item Subgraph coverage check: verifies that the question faithfully covers all relations and properties from the input subgraph, with no omissions, additions, or distortions.
    \item Answer satisfaction check: a search agent verifies that the ground-truth answer indeed satisfies all conditions stated in the question.
\end{itemize}

All LLM-based steps in this stage are performed using DeepSeek-V3.2~\citep{deepseekai2025deepseekv32pushingfrontieropen}.

\subsection{Post Filtering and Human Review}
After the QA generation stage, we subject all questions to multiple rounds of filtering:

\paragraph{Uniqueness verification.} Although the subgraph sampling stage enforces structural uniqueness, the transformation from subgraph to natural language may introduce ambiguity. We deploy multiple search agents of different capability levels to independently attempt each question, collect candidate answers, and automatically judge whether any candidate satisfies all conditions. Questions for which an alternative valid answer is found are filtered out.

\paragraph{Difficulty filtering.} To calibrate the difficulty of the benchmark, we have a DeepSeek-V3.2-powered search agent attempt each question multiple times independently. Questions that are answered correctly across multiple trials are filtered out, retaining only those that pose genuine search difficulty.

\paragraph{Human review.} After all automated filtering, the remaining questions are submitted to professional annotators for manual verification. Annotators evaluate each question along multiple dimensions such as answer correctness, answer uniqueness, logical coherence across conditions, language fluency, and information redundancy, ensuring that the final questions are both rigorous and natural.

\subsection{Data Statistics}

\begin{figure}[t]
    \centering
    \includegraphics[width=\linewidth]{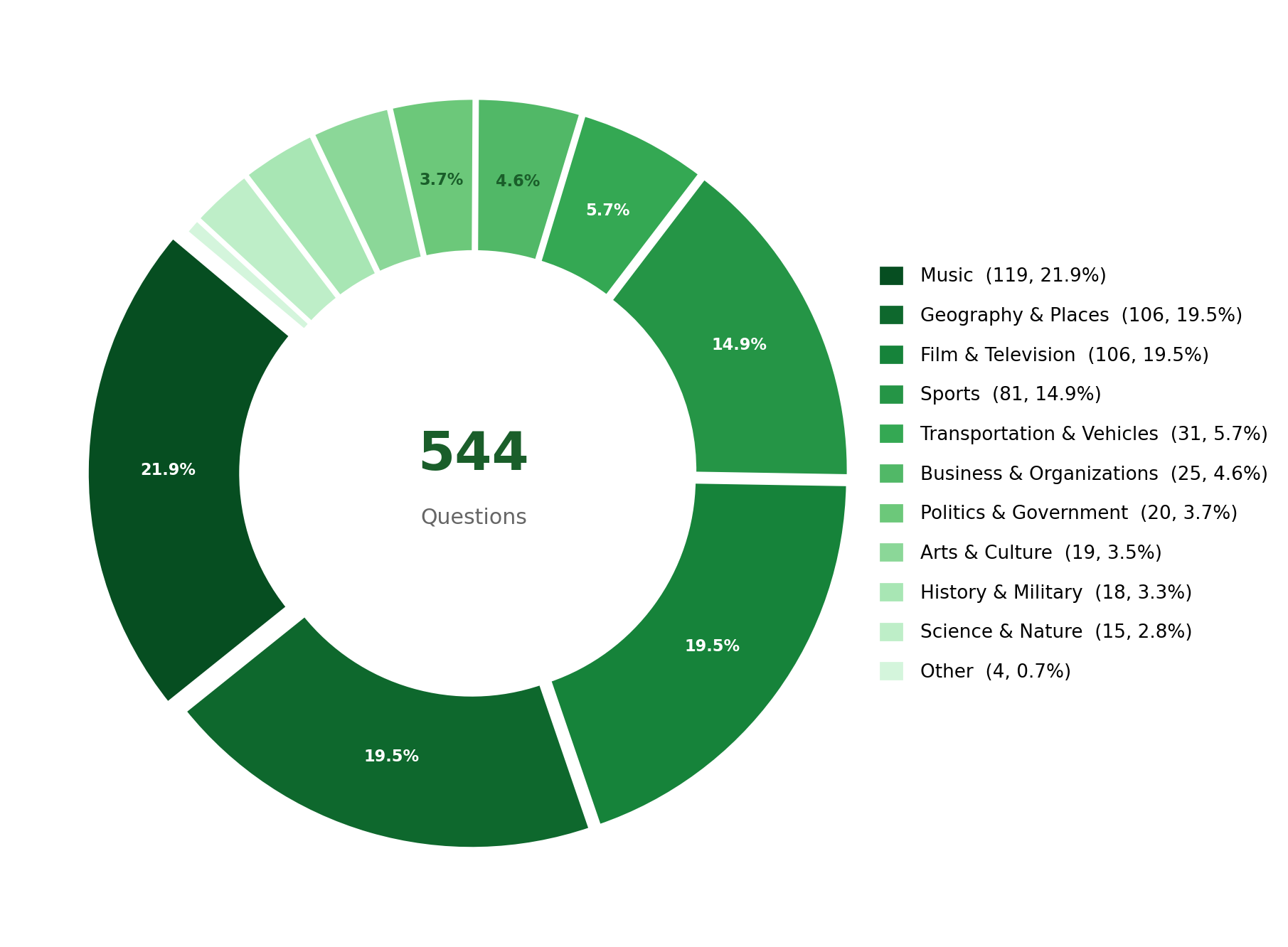}
    \caption{Domain distribution of LoHoSearch. The dataset consists of 544 samples spanning 11 categories.}
    \label{fig:domain_distribution}
\end{figure}

Table~\ref{tab:dataset_stats} summarizes the key statistics of LoHoSearch. The dataset comprises 544 human-verified questions. Graph-structured subgraphs are notably denser than tree-structured ones, with more nodes and nearly twice as many edges, reflecting their higher structural complexity. As shown in Figure~\ref{fig:domain_distribution}, the questions span 11 topical domains, including Music, Geography \& Places, Film \& Television, and Sports, ensuring broad coverage across knowledge areas.

\begin{table}[t]
\centering
\small
\begin{tabular}{lccc}
\toprule
& \textbf{Tree} & \textbf{Graph} & \textbf{Total} \\
\midrule
Questions & 282 & 262 & 544 \\
Avg. Nodes & 7.9 & 10.0 & 8.9 \\
Avg. Edges & 6.9 & 12.5 & 9.6 \\
\bottomrule
\end{tabular}
\caption{Dataset statistics of LoHoSearch.}
\label{tab:dataset_stats}
\end{table}

\begin{table*}[t]
\setlength{\tabcolsep}{4pt}
\centering
\caption{Performance (\%) on LoHoSearch. 
         Score reports the average correct ratio on the 544-sample dataset.
         Calibration Error measures the confidence calibration of each model.
         Best in \textbf{bold}. 
         $^\dagger$Indicates models that encountered service instability or safety refusals during evaluation.}
\label{tab:LoHosearch_results}
\begin{adjustbox}{max width=\textwidth}
\begin{tabular}{lcccc}
\toprule
\textbf{Model} & \textbf{Reasoning} & \textbf{Source} & \textbf{LoHoSearch Score (\%)} & \textbf{Calibration Error (\%)} \\
\midrule
GPT-5.5 \cite{openai2026gpt55}                             & N & Closed & \textbf{34.74} & 48  \\
DeepSeek-V4-Pro$^\dagger$ \cite{deepseek2026v4}           & Y & Open & 15.99          & 35  \\
Claude-Opus-4.6 \cite{anthropic2026opus46}                    & N & Closed & 15.62          & 31  \\
Kimi-K2.6$^\dagger$ \cite{moonshot2026kimik26}                & Y & Open & 15.53          & 56  \\
Gemini-3.1-Pro  \cite{google2026gemini31pro}            & Y & Closed & 13.32          & 72  \\
GLM-5.1  \cite{zhipu2026glm51}                           & Y & Open  & 12.77          & 58 \\
Claude-Opus-4.7   \cite{anthropic2026opus47}                  & N & Closed & 10.29          & 35  \\
DeepSeek-V4-Flash \cite{deepseek2026v4} 
        & Y & Open & 10.02          & 48  \\
LongCat-Flash-Thinking-2601 \cite{team2026longcat} 
      & Y & Open &  9.74          & 76  \\
MiniMax-M2.7 \cite{minimax2026m27}                       & Y & Open &  2.48          & 59  \\
MiniMax-M2.5  \cite{minimax2026m25}                       & Y & Open &  2.29          & 54  \\
\bottomrule
\end{tabular}
\end{adjustbox}
\end{table*}

Regarding quality assurance, 75.5\% of all automatically constructed questions passed human review directly, 22.3\% were accepted after minor wording adjustments by annotators (e.g., correcting unnatural phrasing or removing redundant qualifiers), and only 2.2\% were discarded due to critical issues such as logical inconsistencies, demonstrating the high generation quality of the automated pipeline. In terms of answer uniqueness, 70.8\% of questions are confirmed by annotators to have a definitively unique answer. For the remaining 29.2\%, annotators were unable to conclusively rule out alternatives, yet found no substitute candidates after thorough search. We note that as question difficulty increases, verifying uniqueness itself becomes extremely challenging even for human annotators, further attesting to the inherent complexity of LoHoSearch questions.

\section{Experiments}
\subsection{Experimental Settings}
We select the best-performing model from each major model family based on their results on BrowseComp, which are then designated as our evaluation targets; the vast majority are the latest releases within their respective families. The full list of evaluated models is detailed in Table~\ref{tab:LoHosearch_results}.

Each model is equipped with two tools: (1)~\texttt{search}, which performs keyword queries using traditional search engines such as Google; and (2)~\texttt{browse}, which retrieves detailed content from one or more specified web pages given their URLs. We adopt the same system prompt as BrowseComp to instruct all models. For models that support both thinking and non-thinking modes or adjustable thinking effort, we adopt their official default settings. To ensure fair comparison across models with potentially different optimal temperature configurations, we uniformly set the temperature to 1.0, which also encourages diverse search strategies. The context window is uniformly set to 200K tokens, with 184K allocated for input and 16K for output.

\paragraph{Evaluation.} We adopt an LLM-based automated evaluation approach. Specifically, we compute accuracy once using the BrowseComp grading prompt with GPT-4.1~\citep{openai2025gpt41} as the judge, and a second time using the SimpleQA~\citep{wei2024measuring} grading prompt with Qwen2.5-32B~\cite{qwen2025qwen25technicalreport} as the judge. The final score is the average of the two accuracy values. We find that a single grading combination may be either overly strict or overly lenient, and averaging across two complementary combinations best approximates the true performance of each model.
\subsection{Main Results}

\begin{table}[t]
\centering
\caption{Ablation study on various context management strategies based on 
         DeepSeek-V4-Flash. Best in \textbf{bold}.}
\label{tab:context_management}
\scalebox{0.85}{
\begin{tabular}{p{3.51cm}cc}
\toprule
\textbf{Strategy} & \textbf{BrowseComp} & \textbf{LoHoSearch} \\
\midrule
Baseline                    & 58.84 & 10.02 \\
\midrule
w/ Summary                  & 63.90  & 11.31 \\
w/ Discard-all              & 67.10 & 12.41 \\
w/ Summary + Verify        & 67.81  & 15.35 \\
w/ Discard-all + Verify    & \textbf{72.87}  & \textbf{16.82} \\
\bottomrule
\end{tabular}
}
\end{table}

As shown in Table~\ref{tab:LoHosearch_results}, GPT-5.5 achieves the highest LoHoSearch Score of 34.74\%, substantially outperforming all other models. DeepSeek-V4-Pro, Claude-Opus-4.6, and Kimi-K2.6 demonstrate comparable performance with scores ranging from 15.53\% to 15.99\%, while the remaining models all score below 14\%. Overall, no model exceeds 35\% accuracy, confirming that LoHoSearch poses a substantial challenge to current state-of-the-art search agents. During evaluation, we observed that Kimi-K2.6 and DeepSeek-V4-Pro encountered safety refusals and service instability issues, which may have affected their scores; results impacted by these issues are denoted with $^\dagger$ in Table~\ref{tab:LoHosearch_results}.

Regarding confidence calibration, we adopt the calibration error metric following the methodology of BrowseComp-ZH~\citep{browsecompzh}. We note that some models did not strictly follow the system-prompt format (e.g., omitting the Confidence field), introducing noise into the calibration results. Despite this, models evaluated on our benchmark exhibit consistently high calibration errors, indicating considerable uncertainty when confronted with these challenging queries. This further corroborates the difficulty of LoHoSearch.

\subsection{Context Management}
To investigate how existing context management strategies perform under the extended reasoning demands of LoHoSearch, we conduct experiments on DeepSeek-V4-Flash with a standard ReAct framework~\citep{yao2022react} as the baseline. We then evaluate two context management strategies triggered when token usage exceeds 80\% of the context window: (1) Summary~\citep{wu2025resum}, which compresses the current trajectory and re-initiates the search. (2) Discard-all~\citep{deepseekai2025deepseekv32pushingfrontieropen}, which discards all previous tool calls and restarts with only the original query. Additionally, we incorporate a Verify module that leverages the query and the most recent reasoning steps to judge whether all query conditions are satisfied before committing to a final answer.

As shown in Table~\ref{tab:context_management}, the best-performing combination (Discard-all + Verify) achieves 16.82\%, an absolute gain of only 6.8\% over the baseline. Notably, the same strategies yield a 14.03\% gain on BrowseComp, yet deliver only marginal improvements on LoHoSearch. This discrepancy suggests that the large search space and high structural complexity of our benchmark demand substantially longer reasoning chains, exposing fundamental limitations in current context management approaches. These results also suggest that our benchmark provides a more rigorous testbed for the development of next-generation context management techniques.

\begin{figure}[t]
    \centering
    \includegraphics[width=\linewidth]{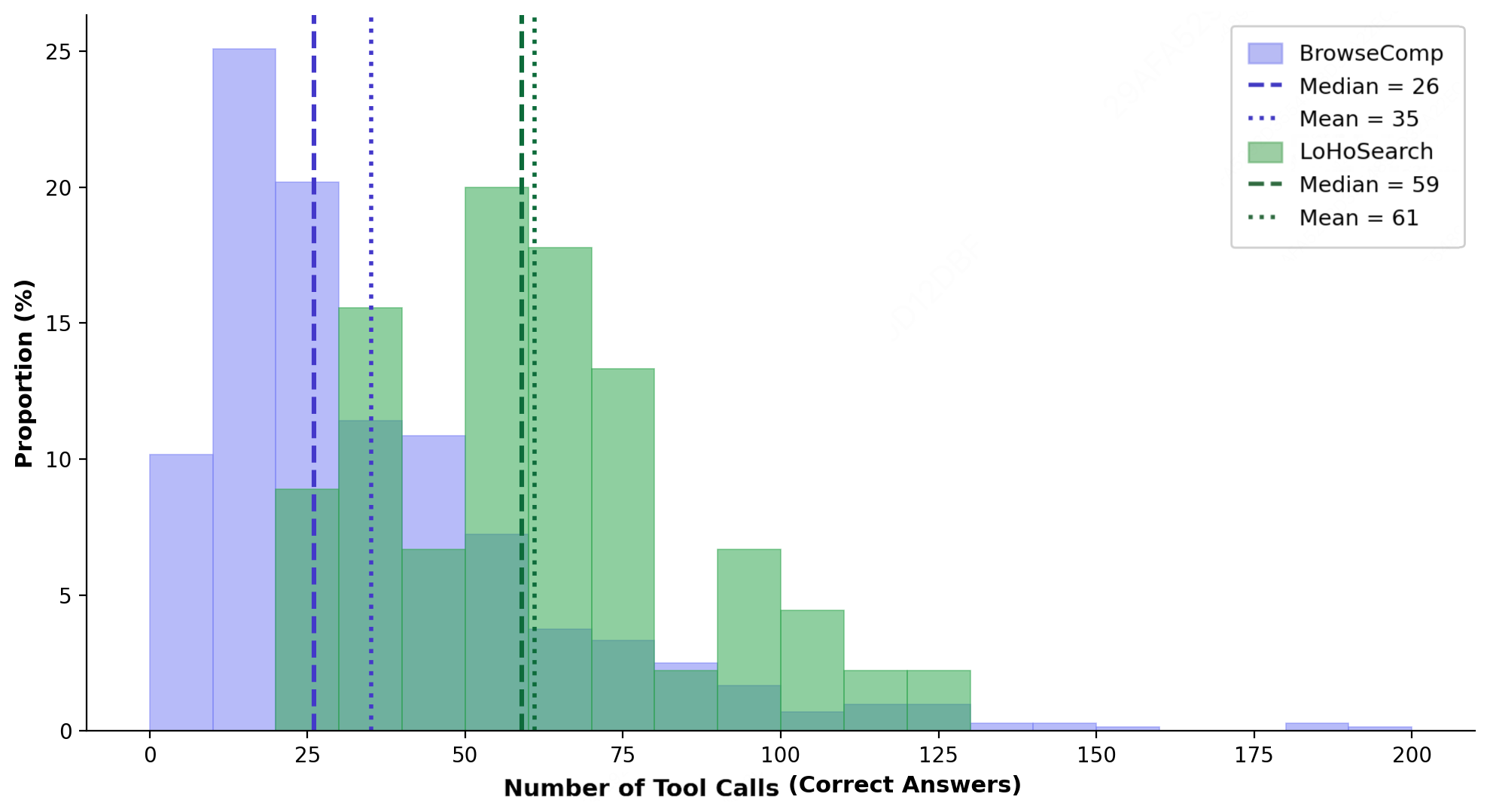}
    \caption{Distribution of the number of tool calls (correct answers) 
             for BrowseComp and LoHoSearch. LoHoSearch requires significantly more tool calls to reach correct answers compared to BrowseComp.}
    \label{fig:rounds_distribution}
\end{figure}

\subsection{Further Analyses}
\subsubsection{Difficulty Analysis}
We further analyze the difficulty gap between BrowseComp and LoHoSearch using DeepSeek-V4-Flash as the probe model. In terms of accuracy, DeepSeek-V4-Flash achieves 58.84\% on BrowseComp but only 10.02\% on LoHoSearch (Table~\ref{tab:LoHosearch_results}). Regarding the distribution of tool calls in correct trajectories, as illustrated in Figure~\ref{fig:rounds_distribution}, solving queries in our benchmark requires substantially more tool calls compared to BrowseComp. Specifically, the mean number of tool calls increases from 35 to 61 (a 74\% relative increase), while the median rises from 26 to 59, confirming that LoHoSearch demands substantially more multi-step reasoning and retrieval.

Furthermore, graph-structured questions achieve an accuracy of only 8.01\%, compared to 11.89\% for tree-structured ones. This gap is primarily attributable to the cyclic dependencies and cross-constraints in graph structures that prevent problem decomposition, indicating that structural complexity, as a difficulty factor independent of search space size, further increases solving difficulty.

\begin{figure}[t]
    \centering
    \includegraphics[width=\linewidth]{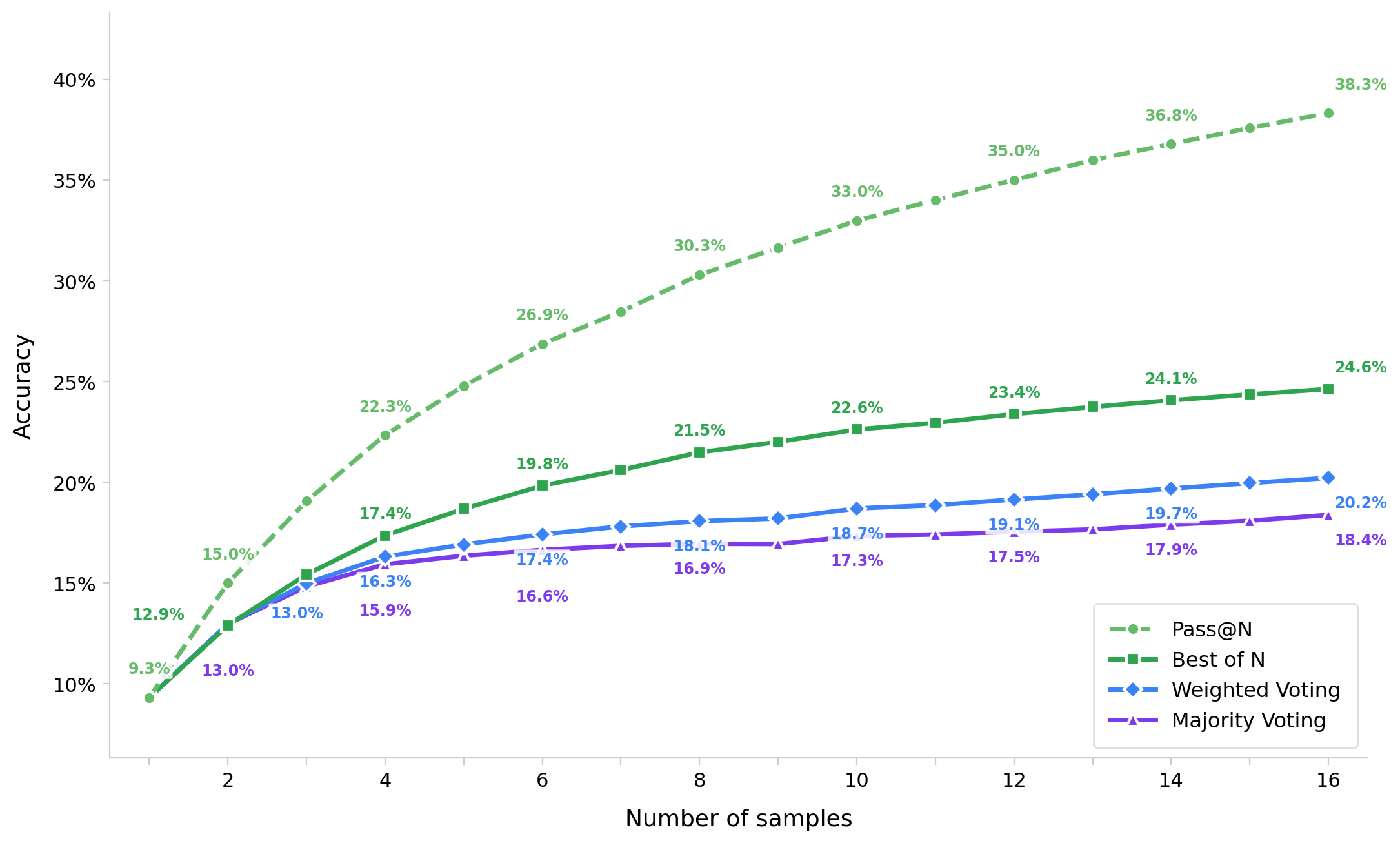}
    \caption{Illustration of pass@N and three answer 
aggregation strategies (majority voting, weighted voting, and 
best-of-N) under parallel sampling.}
    \label{fig:parallel_sampling_voting}
\end{figure}

\subsubsection{Parallel Sampling}

To investigate the performance upper bound achievable through repeated sampling on our benchmark, we conduct a parallel sampling analysis following the scoring formulation of BrowseComp. Specifically, we sample 16 independent responses from DeepSeek-V4-Flash and evaluate three answer aggregation strategies: majority voting, weighted voting, and best-of-N selection.

\begin{figure*}[t]
    \centering
    \includegraphics[width=\linewidth]{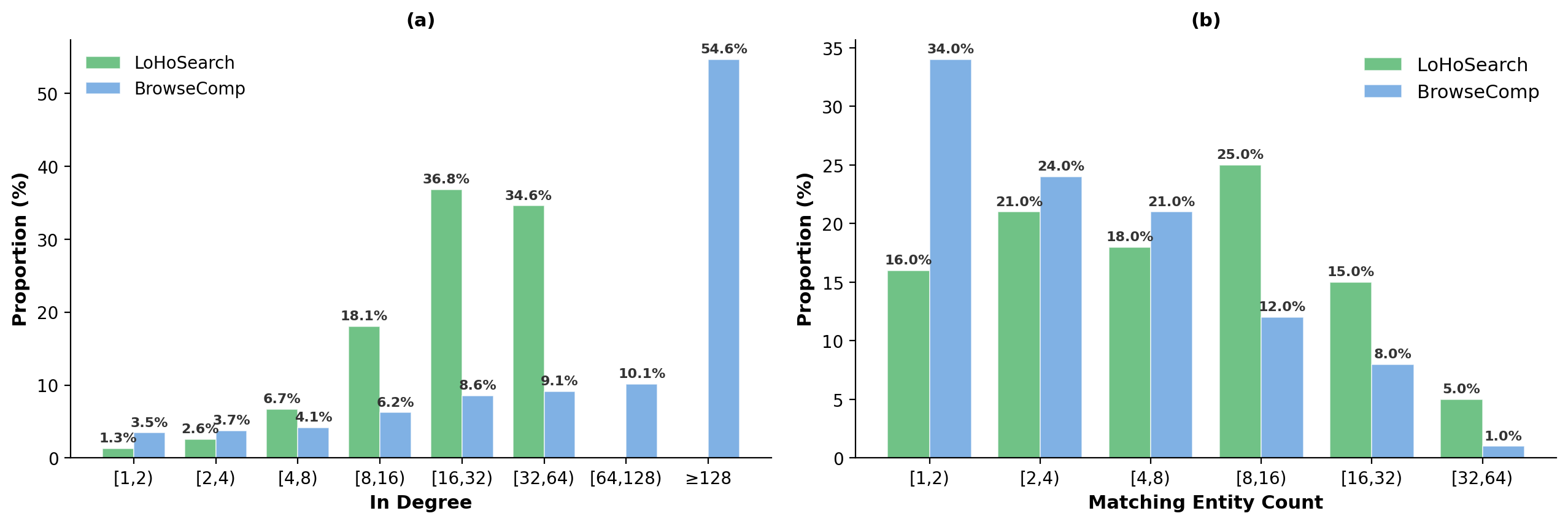}
    \caption{
    Analysis of hidden entity popularity and search space size. (a) In-degree distribution of hidden entities in BrowseComp and LoHoSearch, where higher in-degree indicates greater entity popularity and easier inferability. BrowseComp contains significantly more high-popularity hidden entities than LoHoSearch. (b) Under the same popularity constraint, LoHoSearch exhibits a substantially larger relational search space, demonstrating the necessity of a knowledge graph for systematically constructing high-difficulty questions.
    }
    \label{fig:indegree_analysis}
\end{figure*}

As shown in Figure~\ref{fig:parallel_sampling_voting}, the pass@N metric improves substantially with increasing sample count, rising from 9.3\% at $N{=}1$ to 38.3\% at $N{=}16$, indicating considerable headroom for improvement through repeated sampling. Among the three aggregation strategies, best-of-N, which selects the answer with the highest model confidence score, achieves the strongest performance at 24.6\%. This demonstrates that confidence-based selection is more effective than voting-based approaches and yields performance closer to the theoretical pass@N upper bound.

\subsubsection{Necessity of the Knowledge Graph}

We analyze the hidden entities in questions from BrowseComp and LoHoSearch, i.e., entities that the agent must infer on its own during the solving process, to demonstrate the necessity of a knowledge graph for constructing truly challenging questions. Specifically, we focus on the hidden entities directly associated with the answer in each question. For LoHoSearch, hidden entities are obtained directly from the subgraph structure data; since human-authored benchmarks inherently lack such structural annotations, we parse hidden entities from search agents' correct trajectories, and for questions that agents failed to answer correctly, we use DeepSeek-V3.2 to reverse-engineer them from the answer and question text. Both extraction methods share the same objective: identifying intermediate entities that require reasoning during the solving process. After obtaining the hidden entities, we further retrieve their associated information in the knowledge graph, including entity popularity and the search space size of relations.

The results are shown in Figure~\ref{fig:indegree_analysis}. First, the popularity of hidden entities in BrowseComp is substantially higher than in LoHoSearch (Figure~\ref{fig:indegree_analysis}a). Although BrowseComp also aims to construct challenging questions, the lack of a global perspective in human annotation makes it difficult to precisely control entity popularity, resulting in hidden entities that tend to be relatively easy to infer. Second, even under the same popularity constraint, the relational search space size of entities in LoHoSearch is significantly larger than in BrowseComp (Figure~\ref{fig:indegree_analysis}b), meaning that inferring new entities from a given entity is considerably harder. The above analysis shows that human construction has clear limitations in both entity popularity and search space size, and that a knowledge graph is a necessary foundation for systematically constructing high-difficulty questions.

\section{Related Work}
\subsection{Search Agent Benchmarks}
Multi-hop question answering benchmarks established the foundation for evaluating complex reasoning over documents. HotpotQA~\citep{yang-etal-2018-hotpotqa} contains 113K Wikipedia-based questions covering bridge and comparison reasoning. It was the first dataset to require sentence-level supporting evidence. Building on this, 2WikiMultiHopQA~\citep{ho-etal-2020-constructing} extended coverage to four reasoning types across two Wikipedia sources with chain-level evidence annotations. MuSiQue~\citep{trivedi-etal-2022-musique} composed single-hop sub-questions bottom-up into 25K questions of 2--4 hops and used unanswerable variants to resist shortcuts. These benchmarks share two constraints: they operate in closed-domain settings that do not reflect open-web agent use, and they rely on manual construction that limits scale.

Tool-augmented language models shifted evaluation toward the open web. GAIA~\citep{mialon2024gaia} contains 466 questions at three difficulty levels. FRAMES~\citep{krishna-etal-2025-fact} provides 824 multi-hop questions for RAG pipelines. These benchmarks bring tool use into scope but still rely on fixed, manually curated question sets. They cannot scale beyond annotation budgets, and they cannot be refreshed once published.

BrowseComp~\citep{wei2025browsecomp} serves as the most directly comparable benchmark for our study. It contains 1266 questions designed to be difficult to retrieve but easy to verify, crafted by domain experts through multiple rounds of verification and adversarial filtering. At the time of release, OpenAI Deep Research scored 51.5\%, and human testers 33.3\%. Despite its high quality, its reliance on manual construction introduces clear limitations: (i) construction costs grow linearly with scale; (ii) answer uniqueness relies on human verification without formal guarantees; (iii) difficulty calibration depends on expert judgment. BrowseComp-ZH~\citep{browsecompzh}, which extends the benchmark to Chinese with 289 questions, further illustrates the prohibitive cost of cross-lingual expansion under manual construction.

Recent benchmarks further push difficulty beyond BrowseComp: DeepSearchQA~\citep{deepsearchqa} introduces 900 causal-chain tasks requiring exhaustive answer lists, and WideSearch~\citep{widesearch} evaluates large-scale information collection where even the best system achieves only 5\% success rate. However, both rely on manual construction and frame difficulty through multi-subtask composition rather than maximizing the intrinsic difficulty of individual questions. In contrast, LoHoSearch systematically controls search space size and structural complexity at the single-question level.


In summary, existing benchmarks face three key limitations: (i) manual construction that cannot scale; (ii) answer uniqueness without formal guarantees; (iii) difficulty calibration dependent on human judgment. \textbf{LoHoSearch} addresses all three through a knowledge-graph-driven automated pipeline.

\subsection{QA Generation for Search Agents}
Synthetic data methods for search agents focus on training stronger agents rather than evaluating them. WebShaper~\citep{webshaper} formalizes information seeking into four stages (query formulation, search execution, page browsing, and information synthesis) and repurposes HotpotQA questions to construct agent trajectories. WebSailor~\citep{websailor} introduces a multi-round progressive search framework and achieves superhuman performance on BrowseComp. WebSailor-V2~\citep{websailor-v2} combines synthetic data with scalable reinforcement learning under a progressive difficulty curriculum, significantly narrowing the gap between open-source and proprietary agents.

These works focus on leveraging existing QA datasets as training signals to improve agent performance. In contrast, LoHoSearch takes a fundamentally different approach: We build a large-scale knowledge graph and sample structurally complex subgraphs for QA synthesis. Because the difficulty can be systematically escalated by adjusting search space size and structural complexity parameters, the benchmark keeps pace with the rapid self-improvement cycle of modern agents rather than being left behind by it.

Knowledge-graph-based QA generation constitutes a closely related line of research. KNIGHT~\citep{knight} builds topic-specific knowledge graphs as compressed, reusable representations and generates multiple-choice questions at varying difficulty levels through graph traversal. GraphGen~\citep{GraphGenacl2026} builds fine-grained knowledge graphs from source text and generates QA pairs at multiple granularities via multi-hop neighborhood sampling.

LoHoSearch shares the knowledge-graph-driven generation philosophy with these works but differs in two key respects. First, LoHoSearch uses the graph primarily to guarantee answer validity through subgraph structural uniqueness. Second, rather than probing static knowledge, LoHoSearch prompts an LLM to extract obfuscated relation descriptions from Wikipedia and applies a search verification step to ensure that no question can be resolved through simple retrieval. Together, these choices decouple the knowledge graph from question content and answer storage, extending KG-driven evaluation from closed-domain knowledge assessment to open-web browsing agent evaluation.

\section{Conclusion}
We present LoHoSearch, a benchmark constructed via a knowledge-graph-driven pipeline that systematically controls search space size and structural complexity. The resulting questions surpass the difficulty ceiling of human annotation, and experiments confirm that all evaluated models struggle while current context management strategies yield limited gains. LoHoSearch provides a discriminative testbed for advancing search agent capabilities.

\section*{Limitations}
LoHoSearch currently covers only English. Since our pipeline is language-agnostic by design, we plan to release multilingual variants in future work. Additionally, the released evaluation set is a static snapshot that may become susceptible to contamination or temporal drift over time. On the evaluation side, all models are assessed under the same search-and-browse toolset with a fixed context window, and answer correctness relies on LLM-based judges that may introduce noise on ambiguous edge cases.

Beyond these aspects, two further limitations stem from the construction pipeline itself. First, answer uniqueness is verified within our knowledge graph; for the 29.2\% of questions that human annotators could not conclusively confirm as uniquely answerable (\S2.4), alternative answers outside the KG cannot be ruled out, though no substitute candidate was found after thorough search. Second, difficulty filtering relies on a single LLM (DeepSeek-V3.2) as the calibration agent, which may introduce family-specific bias in which questions survive filtering. We plan to address these aspects through periodic regeneration, broader tool settings, more diverse calibration agents, and improved judging in subsequent iterations.

\section*{Ethical considerations}
This research is conducted in full accordance with the EMNLP Code of Ethics. All datasets used in this work were acquired by strictly following their respective licensing agreements and usage guidelines, with no violation of privacy or data protection standards. We have made deliberate efforts throughout the study to examine and mitigate any potential biases or discriminatory effects that may arise in both the data collection and model development processes. No personally identifiable information was included in any part of our experiments, thereby ensuring the protection of individual privacy and security. We are committed to maintaining transparency, reproducibility, and academic integrity across all stages of this research.

\bibliography{custom}

@inproceedings{yang-etal-2018-hotpotqa,
    title = "{H}otpot{QA}: A Dataset for Diverse, Explainable Multi-hop Question Answering",
    author = "Yang, Zhilin  and
      Qi, Peng  and
      Zhang, Saizheng  and
      Bengio, Yoshua  and
      Cohen, William  and
      Salakhutdinov, Ruslan  and
      Manning, Christopher D.",
    editor = "Riloff, Ellen  and
      Chiang, David  and
      Hockenmaier, Julia  and
      Tsujii, Jun{'}ichi",
    booktitle = "Proceedings of the 2018 Conference on Empirical Methods in Natural Language Processing",
    month = oct # "-" # nov,
    year = "2018",
    address = "Brussels, Belgium",
    publisher = "Association for Computational Linguistics",
    url = "https://aclanthology.org/D18-1259/",
    doi = "10.18653/v1/D18-1259",
    pages = "2369--2380",
}

@inproceedings{ho-etal-2020-constructing,
    title = "Constructing A Multi-hop {QA} Dataset for Comprehensive Evaluation of Reasoning Steps",
    author = "Ho, Xanh  and
      Duong Nguyen, Anh-Khoa  and
      Sugawara, Saku  and
      Aizawa, Akiko",
    editor = "Scott, Donia  and
      Bel, Nuria  and
      Zong, Chengqing",
    booktitle = "Proceedings of the 28th International Conference on Computational Linguistics",
    month = dec,
    year = "2020",
    address = "Barcelona, Spain (Online)",
    publisher = "International Committee on Computational Linguistics",
    url = "https://aclanthology.org/2020.coling-main.580/",
    doi = "10.18653/v1/2020.coling-main.580",
    pages = "6609--6625"
}

@article{wei2025browsecomp,
  title={Browse{C}omp: A simple yet challenging benchmark for browsing agents},
  author={Wei, Jason and Sun, Zhiqing and Papay, Spencer and McKinney, Scott and Han, Jeffrey and Fulford, Isa and Chung, Hyung Won and Passos, Alex Tachard and Fedus, William and Glaese, Amelia},
  journal={arXiv preprint arXiv:2504.12516},
  year={2025}
}

@article{trivedi-etal-2022-musique,
    title = "{M}u{S}i{Q}ue: Multihop Questions via Single-hop Question Composition",
    author = "Trivedi, Harsh  and
      Balasubramanian, Niranjan  and
      Khot, Tushar  and
      Sabharwal, Ashish",
    editor = "Roark, Brian  and
      Nenkova, Ani",
    journal = "Transactions of the Association for Computational Linguistics",
    volume = "10",
    year = "2022",
    address = "Cambridge, MA",
    publisher = "MIT Press",
    url = "https://aclanthology.org/2022.tacl-1.31/",
    doi = "10.1162/tacl_a_00475",
    pages = "539--554"
}

@misc{browsecompzh,
      title={Browse{C}omp-{ZH}: Benchmarking Web Browsing Ability of Large Language Models in Chinese}, 
      author={Peilin Zhou and Bruce Leon and Xiang Ying and Can Zhang and Yifan Shao and Qichen Ye and Dading Chong and Zhiling Jin and Chenxuan Xie and Meng Cao and Yuxin Gu and Sixin Hong and Jing Ren and Jian Chen and Chao Liu and Yining Hua},
      year={2025},
      eprint={2504.19314},
      archivePrefix={arXiv},
      primaryClass={cs.CL},
      url={https://arxiv.org/abs/2504.19314}, 
}

@inproceedings{webshaper,
title={Web{S}haper: Agentically Data Synthesizing via Information-Seeking Formalization},
author={Zhengwei Tao and Jialong Wu and Wenbiao Yin and Pu Wu and Junkai Zhang and Baixuan Li and Haiyang SHEN and Kuan Li and Liwen Zhang and Xinyu Wang and Wentao Zhang and Yong Jiang and Pengjun Xie and Fei Huang and Jingren Zhou},
booktitle={The Fourteenth International Conference on Learning Representations},
year={2026},
url={https://openreview.net/forum?id=hld4TzJsnD}
}

@misc{websailor,
      title={Web{S}ailor: Navigating Super-human Reasoning for Web Agent}, 
      author={Kuan Li and Zhongwang Zhang and Huifeng Yin and Liwen Zhang and Litu Ou and Jialong Wu and Wenbiao Yin and Baixuan Li and Zhengwei Tao and Xinyu Wang and Weizhou Shen and Junkai Zhang and Dingchu Zhang and Xixi Wu and Yong Jiang and Ming Yan and Pengjun Xie and Fei Huang and Jingren Zhou},
      year={2025},
      eprint={2507.02592},
      archivePrefix={arXiv},
      primaryClass={cs.CL},
      url={https://arxiv.org/abs/2507.02592}, 
}

@inproceedings{websailor-v2,
title={Web{S}ailor-{V2}: Bridging the Chasm to Proprietary Agents via Synthetic Data and Scalable Reinforcement Learning},
author={Kuan Li and Zhongwang Zhang and Huifeng Yin and Rui Ye and Yida Zhao and Liwen Zhang and Litu Ou and Ding-Chu Zhang and Xixi Wu and Xinmiao Yu and Jialong Wu and Xinyu Wang and Zile Qiao and Zhen Zhang and Yong Jiang and Pengjun Xie and Fei Huang and Zhi-Qin John Xu and Shuai Wang and Minhao Cheng and Jingren Zhou},
booktitle={The Fourteenth International Conference on Learning Representations},
year={2026},
url={https://openreview.net/forum?id=HuP16O5SJf}
}

@inproceedings{knight,
  title = {{KNIGHT}: Knowledge Graph-Driven Multiple-Choice Question Generation with Adaptive Hardness Calibration},
  author = {Amanlou, Mohammad and {Shafiee Moghaddam}, Erfan and Nouri, Mahdi and {Amou Jafary}, Yasaman and Farsi, Farhan and Bahrak, Behnam},
  booktitle = {Proceedings of the Conference on Parsing and Linguistic Theories (CPAL)},
  year = {2026},
  url = {https://openreview.net/forum?id=8kA9oO5gEc},
}

@misc{GraphGenacl2026,
      title={Graph{G}en: Enhancing Supervised Fine-Tuning for LLMs with Knowledge-Driven Synthetic Data Generation}, 
      author={Zihong Chen and Wanli Jiang and Jinzhe Li and Zhonghang Yuan and Huanjun Kong and Wanli Ouyang and Nanqing Dong},
      year={2025},
      eprint={2505.20416},
      archivePrefix={arXiv},
      primaryClass={cs.CL},
      url={https://arxiv.org/abs/2505.20416}, 
}

@inproceedings{
mialon2024gaia,
title={{GAIA}: a benchmark for General {AI} Assistants},
author={Gr{\'e}goire Mialon and Cl{\'e}mentine Fourrier and Thomas Wolf and Yann LeCun and Thomas Scialom},
booktitle={The Twelfth International Conference on Learning Representations},
year={2024},
url={https://openreview.net/forum?id=fibxvahvs3}
}

@inproceedings{krishna-etal-2025-fact,
    title = "Fact, Fetch, and Reason: A Unified Evaluation of Retrieval-Augmented Generation",
    author = "Krishna, Satyapriya  and
      Krishna, Kalpesh  and
      Mohananey, Anhad  and
      Schwarcz, Steven  and
      Stambler, Adam  and
      Upadhyay, Shyam  and
      Faruqui, Manaal",
    editor = "Chiruzzo, Luis  and
      Ritter, Alan  and
      Wang, Lu",
    booktitle = "Proceedings of the 2025 Conference of the Nations of the Americas Chapter of the Association for Computational Linguistics: Human Language Technologies (Volume 1: Long Papers)",
    month = apr,
    year = "2025",
    address = "Albuquerque, New Mexico",
    publisher = "Association for Computational Linguistics",
    url = "https://aclanthology.org/2025.naacl-long.243/",
    doi = "10.18653/v1/2025.naacl-long.243",
    pages = "4745--4759",
    ISBN = "979-8-89176-189-6"
}

@misc{deepseek2026v4,
  title   = {{DeepSeek-V4}: Towards Highly Efficient Million-Token Context Intelligence},
  author  = {DeepSeek-AI},
  year    = {2026},
  url     = {https://huggingface.co/deepseek-ai/DeepSeek-V4-Pro/blob/main/DeepSeek_V4.pdf},
}

@article{yao2022react,
  title={React: Synergizing reasoning and acting in language models},
  author={Yao, Shunyu and Zhao, Jeffrey and Yu, Dian and Du, Nan and Shafran, Izhak and Narasimhan, Karthik and Cao, Yuan},
  journal={arXiv preprint arXiv:2210.03629},
  year={2022}
}

@article{wu2025resum,
  title={Resum: Unlocking long-horizon search intelligence via context summarization},
  author={Wu, Xixi and Li, Kuan and Zhao, Yida and Zhang, Liwen and Ou, Litu and Yin, Huifeng and Zhang, Zhongwang and Yu, Xinmiao and Zhang, Dingchu and Jiang, Yong and others},
  journal={arXiv preprint arXiv:2509.13313},
  year={2025}
}

@article{team2026longcat,
  title={Long{C}at-{F}lash-{T}hinking-2601 technical report},
  author={LongCat-Team and Gui, Anchun and Li, Bei and Tao, Bingyang and Zhou, Bole and Chen, Borun and Zhang, Chao and Gao, Chen and Zhang, Chen and Han, Chengcheng and others},
  journal={arXiv preprint arXiv:2601.16725},
  year={2026}
}

@misc{anthropic2026claudemythos,
  title   = {System Card: {Claude Mythos Preview}},
  author  = {Anthropic},
  year    = {2026},
  month   = apr,
  url     = {https://www-cdn.anthropic.com/8b8380204f74670be75e81c820ca8dda846ab289.pdf},
}

@misc{anthropic2026opus46,
  title   = {System Card: {Claude Opus 4.6}},
  author  = {Anthropic},
  year    = {2026},
  url     = {https://www-cdn.anthropic.com/6a5fa276ac68b9aeb0c8b6af5fa36326e0e166dd.pdf},
}

@misc{anthropic2026opus47,
  title   = {System Card: {Claude Opus 4.7}},
  author  = {Anthropic},
  year    = {2026},
  url     = {https://cdn.sanity.io/files/4zrzovbb/website/037f06850df7fbe871e206dad004c3db5fd50340.pdf},
}

@misc{openai2026gpt55,
  title   = {Introducing {GPT-5.5}},
  author  = {OpenAI},
  year    = {2026},
  month   = apr,
  url     = {https://openai.com/index/introducing-gpt-5-5/},
}

@misc{zhipu2026glm51,
  title   = {{GLM-5.1}: Towards Long-Horizon Tasks},
  author  = {Zhipu-AI},
  year    = {2026},
  url     = {https://z.ai/blog/glm-5.1},
}

@misc{google2026gemini31pro,
  title   = {Model Card: {Gemini 3.1 Pro}},
  author  = {Google DeepMind},
  year    = {2026},
  url     = {https://storage.googleapis.com/deepmind-media/Model-Cards/Gemini-3-1-Pro-Model-Card.pdf},
}

@misc{moonshot2026kimik26,
  title   = {{Kimi K2.6}: Advancing Open-Source Coding},
  author  = {Moonshot-AI},
  year    = {2026},
  url     = {https://www.kimi.com/blog/kimi-k2-6},
}

@misc{minimax2026m25,
  title   = {{MiniMax M2.5}: Built for Real-World Productivity},
  author  = {MiniMax},
  year    = {2026},
  url     = {https://www.minimaxi.com/news/minimax-m25},
}

@misc{minimax2026m27,
  title   = {{MiniMax M2.7}: Early Echoes of Self-Evolution},
  author  = {MiniMax},
  year    = {2026},
  url     = {https://www.minimax.io/news/minimax-m27-en},
}

@article{wikidata2014denny,
  title	= {Wikidata: A Free Collaborative Knowledge Base},
  author	= {Denny Vrandečić and Markus Krötzsch},
  year	= {2014},
  URL	= {http://cacm.acm.org/magazines/2014/10/178785-wikidata/fulltext},
  journal	= {Communications of the ACM},
  pages	= {78--85},
  volume	= {57}
}

@misc{deepseekai2025deepseekv32pushingfrontieropen,
      title={{DeepSeek-V3.2}: Pushing the Frontier of Open Large Language Models}, 
      author={DeepSeek-AI and Aixin Liu and Aoxue Mei and Bangcai Lin and Bing Xue and Bingxuan Wang and Bingzheng Xu and Bochao Wu and Bowei Zhang and Chaofan Lin and Chen Dong and Chengda Lu and Chenggang Zhao and Chengqi Deng and Chenhao Xu and Chong Ruan and Damai Dai and Daya Guo and Dejian Yang and Deli Chen and Erhang Li and Fangqi Zhou and Fangyun Lin and Fucong Dai and Guangbo Hao and Guanting Chen and Guowei Li and H. Zhang and Hanwei Xu and Hao Li and Haofen Liang and Haoran Wei and Haowei Zhang and Haowen Luo and Haozhe Ji and Honghui Ding and Hongxuan Tang and Huanqi Cao and Huazuo Gao and Hui Qu and Hui Zeng and Jialiang Huang and Jiashi Li and Jiaxin Xu and Jiewen Hu and Jingchang Chen and Jingting Xiang and Jingyang Yuan and Jingyuan Cheng and Jinhua Zhu and Jun Ran and Junguang Jiang and Junjie Qiu and Junlong Li and Junxiao Song and Kai Dong and Kaige Gao and Kang Guan and Kexin Huang and Kexing Zhou and Kezhao Huang and Kuai Yu and Lean Wang and Lecong Zhang and Lei Wang and Liang Zhao and Liangsheng Yin and Lihua Guo and Lingxiao Luo and Linwang Ma and Litong Wang and Liyue Zhang and M. S. Di and M. Y Xu and Mingchuan Zhang and Minghua Zhang and Minghui Tang and Mingxu Zhou and Panpan Huang and Peixin Cong and Peiyi Wang and Qiancheng Wang and Qihao Zhu and Qingyang Li and Qinyu Chen and Qiushi Du and Ruiling Xu and Ruiqi Ge and Ruisong Zhang and Ruizhe Pan and Runji Wang and Runqiu Yin and Runxin Xu and Ruomeng Shen and Ruoyu Zhang and S. H. Liu and Shanghao Lu and Shangyan Zhou and Shanhuang Chen and Shaofei Cai and Shaoyuan Chen and Shengding Hu and Shengyu Liu and Shiqiang Hu and Shirong Ma and Shiyu Wang and Shuiping Yu and Shunfeng Zhou and Shuting Pan and Songyang Zhou and Tao Ni and Tao Yun and Tian Pei and Tian Ye and Tianyuan Yue and Wangding Zeng and Wen Liu and Wenfeng Liang and Wenjie Pang and Wenjing Luo and Wenjun Gao and Wentao Zhang and Xi Gao and Xiangwen Wang and Xiao Bi and Xiaodong Liu and Xiaohan Wang and Xiaokang Chen and Xiaokang Zhang and Xiaotao Nie and Xin Cheng and Xin Liu and Xin Xie and Xingchao Liu and Xingkai Yu and Xingyou Li and Xinyu Yang and Xinyuan Li and Xu Chen and Xuecheng Su and Xuehai Pan and Xuheng Lin and Xuwei Fu and Y. Q. Wang and Yang Zhang and Yanhong Xu and Yanru Ma and Yao Li and Yao Li and Yao Zhao and Yaofeng Sun and Yaohui Wang and Yi Qian and Yi Yu and Yichao Zhang and Yifan Ding and Yifan Shi and Yiliang Xiong and Ying He and Ying Zhou and Yinmin Zhong and Yishi Piao and Yisong Wang and Yixiao Chen and Yixuan Tan and Yixuan Wei and Yiyang Ma and Yiyuan Liu and Yonglun Yang and Yongqiang Guo and Yongtong Wu and Yu Wu and Yuan Cheng and Yuan Ou and Yuanfan Xu and Yuduan Wang and Yue Gong and Yuhan Wu and Yuheng Zou and Yukun Li and Yunfan Xiong and Yuxiang Luo and Yuxiang You and Yuxuan Liu and Yuyang Zhou and Z. F. Wu and Z. Z. Ren and Zehua Zhao and Zehui Ren and Zhangli Sha and Zhe Fu and Zhean Xu and Zhenda Xie and Zhengyan Zhang and Zhewen Hao and Zhibin Gou and Zhicheng Ma and Zhigang Yan and Zhihong Shao and Zhixian Huang and Zhiyu Wu and Zhuoshu Li and Zhuping Zhang and Zian Xu and Zihao Wang and Zihui Gu and Zijia Zhu and Zilin Li and Zipeng Zhang and Ziwei Xie and Ziyi Gao and Zizheng Pan and Zongqing Yao and Bei Feng and Hui Li and J. L. Cai and Jiaqi Ni and Lei Xu and Meng Li and Ning Tian and R. J. Chen and R. L. Jin and S. S. Li and Shuang Zhou and Tianyu Sun and X. Q. Li and Xiangyue Jin and Xiaojin Shen and Xiaosha Chen and Xinnan Song and Xinyi Zhou and Y. X. Zhu and Yanping Huang and Yaohui Li and Yi Zheng and Yuchen Zhu and Yunxian Ma and Zhen Huang and Zhipeng Xu and Zhongyu Zhang and Dongjie Ji and Jian Liang and Jianzhong Guo and Jin Chen and Leyi Xia and Miaojun Wang and Mingming Li and Peng Zhang and Ruyi Chen and Shangmian Sun and Shaoqing Wu and Shengfeng Ye and T. Wang and W. L. Xiao and Wei An and Xianzu Wang and Xiaowen Sun and Xiaoxiang Wang and Ying Tang and Yukun Zha and Zekai Zhang and Zhe Ju and Zhen Zhang and Zihua Qu},
      year={2025},
      eprint={2512.02556},
      archivePrefix={arXiv},
      primaryClass={cs.CL},
      url={https://arxiv.org/abs/2512.02556}, 
}

@misc{deepsearchqa,
      title={Deep{S}earch{QA}: Bridging the Comprehensiveness Gap for Deep Research Agents}, 
      author={Nikita Gupta and Riju Chatterjee and Lukas Haas and Connie Tao and Andrew Wang and Chang Liu and Hidekazu Oiwa and Elena Gribovskaya and Jan Ackermann and John Blitzer and Sasha Goldshtein and Dipanjan Das},
      year={2026},
      eprint={2601.20975},
      archivePrefix={arXiv},
      primaryClass={cs.CL},
      url={https://arxiv.org/abs/2601.20975}, 
}

@inproceedings{widesearch,
      title={Wide{S}earch: Benchmarking Agentic Broad Info-Seeking}, 
      author={Ryan Wong and Jiawei Wang and Junjie Zhao and Li Chen and Yan Gao and Long Zhang and Xuan Zhou and Zuo Wang and Kai Xiang and Ge Zhang and Wenhao Huang and Yang Wang and Ke Wang},
      booktitle={The Fourteenth International Conference on Learning Representations},
      year={2026},
      url={https://openreview.net/forum?id=Q7YUY7zGkZ}
}

@article{wei2024measuring,
  title={Measuring short-form factuality in large language models},
  author={Wei, Jason and Karina, Nguyen and Chung, Hyung Won and Jiao, Yunxin Joy and Papay, Spencer and Glaese, Amelia and Schulman, John and Fedus, William},
  journal={arXiv preprint arXiv:2411.04368},
  year={2024}
}

@misc{openai2025gpt41,
  title   = {Introducing {GPT‑4.1} in the {API}},
  author  = {OpenAI},
  year    = {2025},
  month   = apr,
  url     = {https://openai.com/index/gpt-4-1/},
}

@misc{qwen2025qwen25technicalreport,
      title={Qwen2.5 Technical Report}, 
      author={Qwen: An Yang and Baosong Yang and Beichen Zhang and Binyuan Hui and Bo Zheng and Bowen Yu and Chengyuan Li and Dayiheng Liu and Fei Huang and Haoran Wei and Huan Lin and Jian Yang and Jianhong Tu and Jianwei Zhang and Jianxin Yang and Jiaxi Yang and Jingren Zhou and Junyang Lin and Kai Dang and Keming Lu and Keqin Bao and Kexin Yang and Le Yu and Mei Li and Mingfeng Xue and Pei Zhang and Qin Zhu and Rui Men and Runji Lin and Tianhao Li and Tianyi Tang and Tingyu Xia and Xingzhang Ren and Xuancheng Ren and Yang Fan and Yang Su and Yichang Zhang and Yu Wan and Yuqiong Liu and Zeyu Cui and Zhenru Zhang and Zihan Qiu},
      year={2025},
      eprint={2412.15115},
      archivePrefix={arXiv},
      primaryClass={cs.CL},
      url={https://arxiv.org/abs/2412.15115}, 
}

\appendix
\section{Calibration Error}
\label{sec:calibration_error}

We adopt Expected Calibration Error (ECE) to measure the alignment between model confidence and actual accuracy. Predicted probabilities are partitioned into five equally spaced bins: $[0, 0.2)$, $[0.2, 0.4)$, $[0.4, 0.6)$, $[0.6, 0.8)$, and $[0.8, 1.0]$. ECE is defined as:

\begin{equation}
\text{ECE} = \sum_{i=1}^{K} \frac{n_i}{N} \left| \text{acc}(i) - \text{conf}(i) \right|
\end{equation}

where $K$ is the number of bins, $N$ is the total number of samples, $n_i$ is the number of samples in the $i$-th bin, $\text{acc}(i)$ is the empirical accuracy of the $i$-th bin, and $\text{conf}(i)$ is the average predicted confidence in the $i$-th bin. A lower ECE indicates better calibration between model confidence and actual performance.

\section{System Prompt}
\label{sec:system_prompt}

We adopt the same system prompt as BrowseComp to instruct all models. The prompt content is as follows:

\begin{center}
\fbox{\parbox{0.9\columnwidth}{\small\ttfamily\raggedright
Your response should be in the following format:\\
Explanation: \{your explanation for your final answer\}\\
Exact Answer: \{your succinct, final answer\}\\
Confidence: \{your confidence score between 0\% and 100\% for your answer\}
}}
\end{center}

\section{Tool Definitions}
\label{sec:tool_definitions}

All models are equipped with two tools: \texttt{search} and \texttt{browse}. Their definitions are as follows:

{\small
\begin{verbatim}
{"type": "function",
 "function": {
   "name": "search",
   "description": "Web search, using traditional
     search engines like google, complex
     questions need to be broken down into
     simple queries",
   "parameters": {
     "type": "object",
     "properties": {
       "query": {
         "type": "array",
         "items": {"type": "string"},
         "description": "List of search queries.
           You can search up to 5 queries at
           the same time."}
       },
     "required": ["query"]
   }
 }
}
\end{verbatim}
}

{\small
\begin{verbatim}
{"type": "function",
 "function": {
   "name": "browse",
   "description": "Retrieve detailed content
     from one or more specified web pages by
     providing their URLs.",
   "parameters": {
     "type": "object",
     "required": ["url"],
     "properties": {
       "url": {
         "type": "array",
         "description": "List of urls. You can
           browse up to 3 webpages at the
           same time.",
         "items": {"type": "string"}
       }
     }
   }
 }
}
\end{verbatim}
}

\section{Case Studies}
\label{sec:case_studies}
We present one tree-structured and one graph-structured example from LoHoSearch.
\begin{center}
\fbox{\parbox{0.9\columnwidth}{\small\ttfamily\raggedright
\textbf{Case 1 (Tree-structured):} Consider the following facts. They all pertain to one album. Which album is it?\\
1. It includes singles from another album (Album A), which was created by a group formed on a reality competition television series that originally aired from the early 2000s to the mid-2010s; additionally, Album A includes a solo song by a person (Person D) who has one child, works as a pop singer and television host, and was born in the late 1970s.\\
2. It includes a track written by a person (Person B) who shares the same name as a person (Person E) who earned an award in the early 2000s; this award has been presented annually since the year 2000 and has an alternative name that includes an acronym for a community service initiative.\\
3. It includes tracks produced by a person (Person C) who studied an academic discipline historically defined by roughly three main methods—individual casework, social group work, and community intervention work—and from the late 20th century, many different methods derived from these three classic methods.\\
\textbf{Answer:} The Best of No Angels
}}
\end{center}

\begin{center}
\fbox{\parbox{0.9\columnwidth}{\small\ttfamily\raggedright
\textbf{Case 2 (Graph-structured):} Who is described by the following conditions?\\
1. This person toured with a musical group (Group C). Group C performs in a music genre (Genre D) that is featured at a music festival (Festival E), where Group C was a performer in the early 2010s. The group also released an album (Album H) in the early 2010s, which includes a musical composition (Composition G) by them. Additionally, Group C performs at a recurring event (Event F), and this person has performed at Event F as well.\\
2. They recorded a comedy album at a recurring event (Event I). Event I featured Group C as a guest of honor in the late 2010s.\\
3. Furthermore, this person appeared as a guest on a podcast (Podcast B) and wrote and performed shows at a film festival (Festival J).\\
\textbf{Answer:} Joseph Scrimshaw
}}
\end{center}

\end{document}